\documentclass[11pt]{article}
\usepackage{emnlp2016}
\usepackage{amsmath}
\usepackage{graphicx}
\usepackage{multirow}
\usepackage{times}
\usepackage{url}
\usepackage{latexsym}
\emnlpfinalcopy

\title{The Effects of Data Size and Frequency Range\\
on Distributional Semantic Models}

\author{Magnus Sahlgren \\
  Gavagai and SICS\\
  Slussplan 9, Box 1263 \\
  111 30 Stockholm, 164 29 Kista \\
  Sweden \\
  {\tt mange@[gavagai|sics].se} \\\And
  Alessandro Lenci \\
  University of Pisa \\
  via Santa Maria 36 \\
  56126 Pisa \\
  Italy \\
  {\tt alessandro.lenci@unipi.it} \\}

\date{}

\begin{document}
\maketitle
\begin{abstract}
This paper investigates the effects of data size and frequency range on distributional semantic models. We compare the performance of a number of representative models for several test settings over data of varying sizes, and over test items of various frequency. Our results show that neural network-based models underperform when the data is small, and that the most reliable model over data of varying sizes and frequency ranges 
is the inverted factorized model.
\end{abstract}

\section{Introduction}

Distributional Semantic Models (DSMs) have become a staple in natural language processing. The various parameters of DSMs --- e.g.~size of context windows, weighting schemes, dimensionality reduction techniques, and similarity measures --- have been thoroughly studied \cite{Weeds:2004,Sahlgren:2006,Riordan:2011,Bullinaria:2012,Levy:2015}, and are now well understood. The impact of various processing models --- matrix-based models, neural networks, and hashing methods --- have also enjoyed considerable attention lately, with at times conflicting conclusions \cite{Baroni:2014,Levy:2015,Schnabel:2015,Osterlund:2015,Sahlgren:2016}. The consensus interpretation of such experiments seems to be that the choice of processing model is less important than the parameterization of the models, since the various processing models all result in more or less equivalent DSMs (provided that the parameterization is comparable).

One of the least researched aspects of DSMs is the effect on the various models of data size and frequency range of the target items. The only previous work in this direction that we are aware of is Asr et al.~\shortcite{Asr:2016}, who report that on small data (the CHILDES corpus), simple matrix-based models outperform neural network-based ones. Unfortunately, Asr et al.~do not include any experiments using the same models applied to bigger data, making it difficult to compare their results with previous studies, since implementational details and parameterization will be different. 

There is thus still a need for a consistent and fair comparison of the performance of various DSMs when applied to data of varying sizes. In this paper, we seek an answer to the question: {\bf which DSM should we opt for if we only have access to limited amounts of data?} We are also interested in the related question: {\bf which DSM should we opt for if our target items are infrequent?} The latter question is particularly crucial, since one of the major assets of DSMs is their applicability to create semantic representations for ever-expanding vocabularies from text feeds, in which new words may continuously appear in the low-frequency ranges. 

In the next section, we introduce the contending DSMs and the general experiment setup, before turning to the experiments and our interpretation of the results. We conclude with some general advice.

\section{Distributional Semantic Models}

One could classify DSMs in many different ways, such as the type of context and the method to build distributional vectors. Since our main goal here is to gain an understanding of the effect of data size and frequency range on the various models, we focus primarily on the differences in processing models, hence the following typology of DSMs.

\subsubsection*{Explicit matrix models}
\noindent
We here include what could be referred to as \emph{explicit} models, in which each vector dimension corresponds to a specific context \cite{Levy:Goldberg:2014}. The baseline model is a simple co-occurrence matrix $F$ (in the following referred to as \textsc{co} for Co-Occurrence). We also include the model that results from applying Positive Pointwise Mutual Information (\textsc{ppmi}) to the co-occurrence matrix. \textsc{ppmi} is defined as simply discarding any negative values of the PMI, computed as:

\begin{equation}\label{eq:ppmi}
\textrm{PMI}(a,b) = \log \frac{f_{ab} \times T}{f_a f_b}
\end{equation}
\noindent
where $f_{ab}$ is the co-occurrence count of word $a$ and word $b$, $f_a$ and $f_b$ are the individual frequencies of the words, and $T$ is the number of tokens in the data.\footnote{We also experimented with {\em smoothed} \textsc{ppmi}, which raises the context counts to the power of $\alpha$ and normalizes them \cite{Levy:2015}, thereby countering the tendency of mutual information to favor infrequent events: $f(b) = \frac{\#(b)^\alpha}{\sum_b \#(b)^\alpha}$, but it did not lead to any consistent improvements compared to PPMI.}

\subsubsection*{Factorized matrix models}
\noindent
This type of model applies an additional factorization of the weighted co-occurrence counts. We here include two variants of applying Singular Value Decomposition (SVD) to the \textsc{ppmi}-weighting co-occurrence matrix; one version that discards all but the first couple of hundred latent dimensions (\textsc{tsvd} for {\em truncated} SVD), and one version that instead {\em removes} the first couple of hundred latent dimensions (\textsc{isvd} for {\em inverted} SVD). SVD is defined in the standard way:

\begin{equation}\label{eq:svd}
F = U\Sigma V^{T}
\end{equation}
where $U$ holds the eigenvectors of $F$, $\Sigma$ holds the eigenvalues, and $V \in U(w)$ is a unitary matrix mapping the original basis of $F$ into its eigenbasis. Since $V$ is redundant due to invariance under unitary transformations, we can represent the factorization of $\hat{F}$ in its most compact form $\hat{F} \equiv U\Sigma$.

\subsubsection*{Hashing models}
\noindent
A different approach to reduce the dimensionality of DSMs is to use a hashing method such as Random Indexing (\textsc{ri}) \cite{Kanerva:2000}, which accumulates distributional vectors $\vec{d}(a)$ in an online fashion:

\begin{equation}\label{eq:ri}
\vec{d}(a) \leftarrow \vec{d}(a_{i}) +  \sum_{j=-c,j\neq 0}^{c} w(x^{(i+j)})\pi^{j}\vec{r}(x^{(i+j)})
\end{equation}
\noindent
where $c$ is the extension of the context window, $w(b)$ is a weight that quantifies the importance of context term $b$,\footnote{We use $w(b) = e^{-{\lambda \cdot \frac{f(b)}{V}}}$ where $f(b)$ is the frequency of context item $b$, $V$ is the total number of unique context items seen thus far (i.e.~the current size of the growing vocabulary), and $\lambda$ is a constant that we set to 60 \cite{Sahlgren:2016}.} $\vec{r}_{d}(b)$ is a {\em sparse random index vector} that acts as a fingerprint of context term $b$, and $\pi^{j}$ is a permutation that rotates the random index vectors one step to the left or right, depending on the position of the context items within the context windows, thus enabling the model to take word order into account \cite{Sahlgren:2008}.

\subsubsection*{Neural network models}
\noindent
There are many variations of DSMs that use neural networks as processing model, ranging from simple recurrent networks \cite{Elman:1990} to more complex deep architectures \cite{Collobert:2008}. The incomparably most popular neural network model is the one implemented in the \texttt{word2vec} library, which uses the softmax for predicting $b$ given $a$ \cite{Mikolov:2013}:

\begin{equation}\label{eq:sgns}
p(b|a) = \frac{\textrm{exp}(\vec{b}\cdot\vec{a})}{\sum_{b'\in C} \textrm{exp}(\vec{b'}\cdot\vec{a})}
\end{equation}
\noindent
where $C$ is the set of context words, and $\vec{b}$ and $\vec{a}$ are the vector representations for the context and target words, respectively. We include two versions of this general model; Continuous Bag of Words (\textsc{cbow}) that predicts a word based on the context, and SkipGram Negative Sampling (\textsc{sgns}) that predicts the context based on the current word.

\section{Experiment setup}

Since our main focus in this paper is the performance of the above-mentioned DSMs on data of varying sizes, we use one big corpus as starting point, and split the data into bins of varying sizes. We opt for the ukWaC corpus \cite{Ferraresi:2008}, which comprises some 1.6 billion words after tokenization and lemmatization. We produce sub-corpora by taking the first 1 million, 10 million, 100 million, and 1 billion words. 

Since the co-occurrence matrix built from the 1 billion-word ukWaC sample is very big (more than 4,000,000 $\times$ 4,000,000), we prune the co-occurrence matrix to 50,000 dimensions before the factorization step by simply removing  infrequent context items.\footnote{Such drastic reduction has a negative effect on the performance of the factorized methods for the 1 billion word data, but unfortunately is necessary for computational reasons.} As comparison, we use 200 dimensions for \textsc{tsvd}, 2,800 (3,000-200) dimensions for \textsc{isvd}, 2,000 dimensions for \textsc{ri}, and 200 dimensions for \textsc{cbow} and \textsc{sgns}. These dimensionalities have been reported to perform well for the respective models \cite{Landauer:1997,Sahlgren:2008,Mikolov:2013,Osterlund:2015}. All DSMs use the same parameters as far as possible with a narrow context window of $\pm 2$ words, which has been shown to produce good results in semantic tasks \cite{Sahlgren:2006,Bullinaria:2012}.

We use five standard benchmark tests in these experiments; two multiple-choice vocabulary tests (the TOEFL synonyms and the ESL synonyms), and three similarity/relatedness rating benchmarks (SimLex-999 (SL) \cite{Hill:2015}, MEN \cite{Bruni:2014}, and Stanford Rare Words (RW) \cite{Luong:2013}). The vocabulary tests measure the synonym relation, while the similarity rating tests measure a broader notion of semantic similarity (SL and RW) or relatedness (MEN).\footnote{It is likely that the results on the similarity tests could be improved by using a wider context window, but such improvement would probably be consistent across all models, and is thus outside the scope of this paper.} The results for the vocabulary tests are given in accuracy (i.e.,~percentage of correct answers), while the results for the similarity tests are given in Spearman rank correlation.

\begin{table}[h!]
{\small
\begin{center}
\begin{tabular}{|l|rr|rrr|}
\hline
{\bf DSM} & {\bf TOEFL} & {\bf ESL} & {\bf SL} & {\bf MEN} & {\bf RW} \\
\hline\hline
\multicolumn{6}{|c|}{1 million words} \\
\hline
\textsc{co} & 17.50 & {\bf 20.00} &  $-$1.64 &  10.72 & $-$3.96 \\
\textsc{ppmi} & 26.25 & 18.00 & 8.28 & 21.49 & $-$2.57 \\
\hline
\textsc{tsvd} & {\bf 27.50} & {\bf 20.00} & 4.43 & {\bf 22.15} & $-$1.56 \\
\textsc{isvd} & 22.50 & 14.00 & {\bf 14.33} & 19.74 & {\bf 5.31} \\
\hline
\textsc{ri} & 20.00 & 16.00 & 5.65 & 17.94 & 1.92 \\
\hline
\textsc{sgns} & 15.00 & 8.00 & 3.64 & 12.34 & 1.46 \\
\textsc{cbow} & 15.00 & 10.00 & $-$0.16 & 11.59 & 1.39 \\
\hline\hline
\multicolumn{6}{|c|}{10 million words} \\
\hline
\textsc{co} & 40.00 & 22.00 & 4.77 & 15.20 & 0.95 \\
\textsc{ppmi} & 52.50 & 38.00 & 26.44 & 39.83 & 4.00 \\
\hline
\textsc{tsvd} & 38.75 & 30.00 & 19.27 & 34.33 & 5.53 \\
\textsc{isvd} & 45.00 & {\bf 44.00} & {\bf 30.19} & {\bf 44.21} & {\bf 9.88} \\
\hline
\textsc{ri} & {\bf 47.50} & 24.00 & 20.44 & 34.56 & 3.32 \\
\hline
\textsc{sgns} & 43.75 & 42.00 & 28.30 & 26.59 & 2.38 \\
\textsc{cbow} & 40.00 & 30.00 & 22.22 & 28.33 & 3.04 \\
\hline\hline
\multicolumn{6}{|c|}{100 million words} \\
\hline
\textsc{co} & 45.00 & 30.00 & 10.00 & 19.36 & 3.12 \\
\textsc{ppmi} & {\bf 66.25} & 54.00 & 33.75 & 46.74 & 15.05 \\
\hline
\textsc{tsvd} & 46.25 & 34.00 & 25.11 & 42.49 & 13.00 \\
\textsc{isvd} & 66.25 & {\bf 66.00} & {\bf 40.98} & {\bf 54.55} & {\bf 21.27} \\
\hline
\textsc{ri} & 55.00 & 48.00 & 32.31 & 45.71 & 10.15 \\
\hline
\textsc{sgns} & 65.00 & 58.00 & 40.75 & 52.83 & 11.73 \\
\textsc{cbow} & 61.25 & 46.00 & 36.15 & 48.30 & 15.62 \\
\hline\hline
\multicolumn{6}{|c|}{1 billion words}\\
\hline
\textsc{co} & 55.00 & 40.00 & 11.85 & 21.83 & 6.82 \\
\textsc{ppmi} & 71.25 & 54.00 & 35.69 & 52.95 & 24.29 \\
\hline
\textsc{tsvd} & 56.25 & 46.00 & 31.36 & 52.05 & 13.35 \\
\textsc{isvd} & 71.25 & {\bf 66.00} & {\bf 44.77} & 60.11 & {\bf 28.46} \\
\hline
\textsc{ri} & 61.25 & 50.00 & 35.35 & 50.51 & 18.58 \\
\hline
\textsc{sgns} & {\bf 76.25} & {\bf 66.00} & 41.94 & {\bf 67.03} & 24.50 \\
\textsc{cbow} & 75.00 & 56.00 & 38.31 & 59.84 & 22.80 \\
\hline
\end{tabular}
\caption{Results for DSMs trained on data of varying sizes.}
\label{tab:size}
\end{center}}
\end{table}

\begin{figure}[h]
\scalebox{.63}{\input{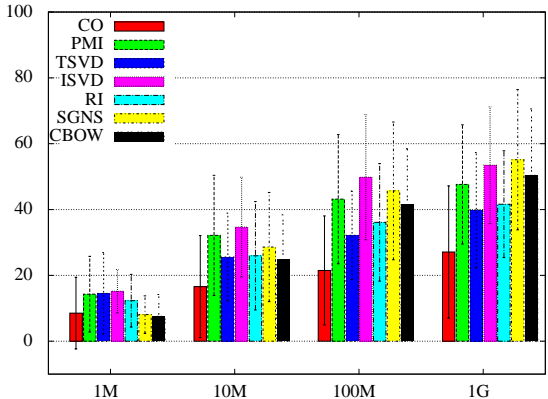}}
\caption{Average results and standard deviation over all tests.}
\label{fig:sizes}
\end{figure}

\section{Comparison by data size}

Table \ref{tab:size} summarizes the results over the different test settings. The most notable aspect of these results is that the neural networks models do not produce competitive results for the smaller data, which corroborates the results by Asr et al.~\shortcite{Asr:2016}. The best results for the smallest data are produced by the factorized models, with both \textsc{tsvd} and \textsc{isvd} producing top scores in different test settings. It should be noted, however, that even the top scores for the smallest data set are substandard; only two models (\textsc{ppmi} and \textsc{tsvd}) manage to beat the random baseline of 25\% for the TOEFL tests, and none of the models manage to beat the random baseline for the ESL test.

The \textsc{isvd} model produces consistently good results; it yields the best overall results for the 10 million and 100 million-word data, and is competitive with \textsc{sgns} on the 1 billion word data. Figure \ref{fig:sizes} shows the average results and their standard deviations over all test settings.\footnote{Although rank correlation is not directly comparable with accuracy, they are both  bounded between zero and one, which means we can take the average to get an idea about overall performance.} It is obvious that there are no huge differences between the various models, with the exception of the baseline \textsc{co} model, which consistently underperforms. The \textsc{tsvd} and \textsc{ri} models have comparable performance across the different data sizes, which is systematically lower than the \textsc{ppmi} model. The \textsc{isvd} model is the most consistently good model, with the neural network-based models steadily improving as data becomes bigger.

Looking at the different datasets, SL and RW are the hardest ones for all the models. In the case of SL, this confirms the results in \cite{Hill:2015}, and might be due to the general bias of DSMs towards semantic relatedness, rather than genuine semantic similarity, as represented in SL. The substandard performance on RW might instead be due to the low frequency of the target items. It is interesting to note that these are benchmark tests in which neural models perform the worst even when trained on the largest data.

\section{Comparison by frequency range}

In order to investigate how each model handles different frequency ranges, we split the test items into three different classes that contain about a third of the frequency mass of the test items each. This split was produced by collecting all test items into a common vocabulary, and then sorting this vocabulary by its frequency in the ukWaC 1 billion-word corpus. We split the vocabulary into 3 equally large parts; the \textsc{HIGH} range with frequencies ranging from 3,515,086 (``do") to 16,830 (``organism"), the \textsc{MEDIUM} range with frequencies ranging between 16,795 (``desirable") and 729 (``prickly"), and the \textsc{LOW} range  with frequencies ranging between 728 (``boardwalk") to hapax legomenon. We then split each individual test into these three ranges, depending on the frequencies of the test items. Test pairs were included in a given frequency class if and only if both the target and its relatum occur in the frequency range for that class. For the constituent words in the test item that belong to different frequency ranges, which is the most common case, we use a separate \textsc{MIXED} class. The resulting four classes contain 1,387 items for the \textsc{HIGH} range, 656 items for the \textsc{MEDIUM} range, 350 items for the \textsc{LOW} range, and 3,458 items for the \textsc{MIXED} range.\footnote{233 test terms did not occur in the 1 billion-word corpus.}

\begin{table*}[t]
\begin{center}
\begin{tabular}{|l|c|c|c||c|}
\hline
{\bf DSM} & {\bf \textsc{HIGH}} & {\bf \textsc{MEDIUM}} & {\bf \textsc{LOW}} & {\bf \textsc{MIXED}} \\
\hline\hline
\textsc{co} & 32.61 ($\uparrow$62.5,$\downarrow$04.6) & 35.77 ($\uparrow$66.6,$\downarrow$21.2) & 12.57 ($\uparrow$35.7,$\downarrow$00.0) & 27.14 ($\uparrow$56.6,$\downarrow$07.9) \\
\textsc{ppmi} & 55.51 ($\uparrow$75.3,$\downarrow$28.0) & 57.83 ($\uparrow$88.8,$\downarrow$18.7) &  25.84 ($\uparrow$50.0,$\downarrow$00.0) & 47.73 ($\uparrow$83.3,$\downarrow$27.1) \\
\hline
\textsc{tsvd} & 50.52 ($\uparrow$70.9,$\downarrow$23.2) & 54.75 ($\uparrow$77.9,$\downarrow$24.1) & 17.85 ($\uparrow$50.0,$\downarrow$00.0) & 41.08 ($\uparrow$56.6,$\downarrow$19.6) \\
\textsc{isvd} & 63.31 ($\uparrow$87.5,$\downarrow$36.5) & {\bf 69.25} ($\uparrow$88.8,$\downarrow$46.3) & 10.94 ($\uparrow$16.0,$\downarrow$00.0) & {\bf 57.24} ($\uparrow$83.3,$\downarrow$33.0) \\
\hline
\textsc{ri} & 53.11 ($\uparrow$62.5,$\downarrow$30.1) & 48.02 ($\uparrow$72.2,$\downarrow$20.4) & 23.29 ($\uparrow$39.0,$\downarrow$00.0) & 46.39 ($\uparrow$66.6,$\downarrow$21.0) \\
\hline
\textsc{sgns} & {\bf 68.81} ($\uparrow$87.5,$\downarrow$36.4) & 62.00 ($\uparrow$83.3,$\downarrow$27.4) & 18.76 ($\uparrow$42.8,$\downarrow$00.0) & 56.93 ($\uparrow$83.3,$\downarrow$30.2) \\
\textsc{cbow} & 62.73 ($\uparrow$81.2,$\downarrow$31.9) & 59.50 ($\uparrow$83.3,$\downarrow$32.4) & {\bf 27.13} ($\uparrow$78.5,$\downarrow$00.0) & 52.21 ($\uparrow$76.6,$\downarrow$25.9) \\
\hline
\end{tabular}
\caption{Average results for DSMs over four different frequency ranges for the items in the TOEFL, ESL, SL, MEN, and RW tests. All DSMs are trained on the 1 billion words data.}
\label{tab:freqs}
\end{center}
\end{table*}

Table \ref{tab:freqs} (next side) shows the average results over the different frequency ranges for the various DSMs trained on the 1 billion-word ukWaC data. We also include the highest and lowest individual test scores (signified by $\uparrow$ and $\downarrow$), in order to get an idea about the consistency of the results. As can be seen in the table, the most consistent model is \textsc{isvd}, which produces the best results in both the \textsc{MEDIUM} and \textsc{MIXED} frequency ranges. The neural network models \textsc{sgns} and \textsc{cbow} produce the best results in the \textsc{HIGH} and \textsc{LOW} range, respectively, with \textsc{cbow} clearly outperforming \textsc{sgns} in the latter case. The major difference between these models is that \textsc{cbow} predicts a word based on a context, while \textsc{sgns} predicts a context based on a word. Clearly, the former approach is more beneficial for low-frequent items.

The \textsc{ppmi}, \textsc{tsvd} and \textsc{ri} models perform similarly across the frequency ranges, with \textsc{ri} producing somewhat lower results in the \textsc{MEDIUM} range, and \textsc{tsvd} producing somewhat lower results in the \textsc{LOW} range. The \textsc{co} model underperforms in all frequency ranges. Worth noting is the fact that all models that are based on an explicit matrix (i.e.~\textsc{co}, \textsc{ppmi}, \textsc{tsvd} and \textsc{isvd}) produce better results in the \textsc{MEDIUM} range than in the \textsc{HIGH} range.

The arguably most interesting results are in the \textsc{LOW} range. Unsurprisingly, there is a general and significant drop in performance for low frequency items, but with interesting differences among the various models. As already mentioned, the \textsc{cbow} model produces the best results, closely followed by \textsc{ppmi} and \textsc{ri}. It is noteworthy that the low-dimensional embeddings of the \textsc{cbow} model only gives a modest improvement over the high-dimensional explicit vectors of \textsc{ppmi}. The worst results are produced by the \textsc{isvd} model, which scores even lower than the baseline \textsc{co} model. This might be explained by the fact that \textsc{isvd} removes the latent dimensions with largest variance, which are arguably the most important dimensions for very low-frequent items. Increasing the number of latent dimensions with high variance in the \textsc{isvd} model improves the results in the \textsc{LOW} range (16.59 when removing only the top 100 dimensions). 

\section{Conclusion}

Our experiments confirm the results of Asr et al.~\shortcite{Asr:2016}, who show that neural network-based models are suboptimal to use for smaller amounts of data. On the other hand, our results also show that none of the standard DSMs work well in situations with small data. It might be an interesting novel research direction to investigate how to design DSMs that are applicable to small-data scenarios. 

Our results demonstrate that the inverted factorized model (\textsc{isvd}) produces the most robust results over data of varying sizes, and across several different test settings. We interpret this finding as further corroborating the results of Bullinaria and Levy \shortcite{Bullinaria:2012}, and \"{O}sterlund et al.~\shortcite{Osterlund:2015}, with the conclusion that the inverted factorized model is a robust competitive alternative to the widely used \textsc{sgns} and \textsc{cbow} neural network-based models.

We have also investigated the performance of the various models on test items in different frequency ranges, and our results in these experiments demonstrate that all tested models perform optimally in the medium-to-high frequency ranges. Interestingly, all models based on explicit count matrices (\textsc{co}, \textsc{ppmi}, \textsc{tsvd} and \textsc{isvd}) produce somewhat better results for items of medium frequency than for items of high frequency. The neural network-based models and \textsc{isvd}, on the other hand, produce the best results for high-frequent items.

None of the tested models perform optimally for low-frequent items. The best results for low-frequent test items in our experiments were produced using the \textsc{cbow} model, the \textsc{ppmi} model and the \textsc{ri} model, all of which uses weighted context items without any explicit factorization. By contrast, the \textsc{isvd} model underperforms significantly for the low-frequent items, which we suggest is an effect of removing latent dimensions with high variance. 

This interpretation suggests that it might be interesting to investigate {\em hybrid models} that use different processing models --- or at least different parameterizations --- for different frequency ranges, and for different data sizes. We leave this as a suggestion for future research.

\section{Acknowledgements}

This research was supported by the Swedish Research Council under contract 2014-28199.

\bibliographystyle{emnlp2016}
\bibliography{main}

\end{document}